\documentclass[11pt]{article}

\usepackage[final]{acl}

\usepackage{times}
\usepackage{latexsym}
\usepackage{amsmath}
\usepackage{algorithm}
\usepackage{amssymb}
\usepackage{multicol, multirow}
\usepackage{booktabs}
\usepackage{array}
\usepackage{makecell}
\usepackage{xcolor}
\usepackage{caption}
\usepackage{tabularx}
\usepackage[T1]{fontenc}

\usepackage[utf8]{inputenc}

\usepackage{microtype}

\usepackage{inconsolata}

\usepackage{graphicx}

\title{Think Before You Comfort: Reflective Cognitive Alignment for Protocol-Grounded Elderly Stimulation Agents}

\author{
\bf Jiyue Jiang$^{\heartsuit}$$^{\ast}$,
Ziyi Li$^{\heartsuit}$$^{\ast}$,
He Hu$^{\clubsuit}$\thanks{These authors are co-first authors.},
Sheng Wang$^{\spadesuit}$,
Yuhan Chen$^{\heartsuit}$,
Yanyu Chen$^{\heartsuit}$,\\
\bf Jingqi Zhou$^{\spadesuit}$,
Pengan Chen$^{\heartsuit}$,
Fei Ma$^{\clubsuit}$, 
Irwin King$^{\heartsuit}$,
Yu Li$^{\heartsuit}$,
Chuan Wu$^{\spadesuit}$\\
$^{\heartsuit}$ The Chinese University of Hong Kong, $^{\spadesuit}$ The University of Hong Kong,\\
$^{\clubsuit}$ Guangdong Provincial Laboratory of Artificial Intelligence and Digital Economy (Shenzhen)\\
{\tt
\{jiangjy, 1155269491, 1155217301, chenyanyu.cse\}@link.cuhk.edu.hk,
} \\
{\tt
\{huhe, mafei\}@gml.ac.cn,
\{u3638070, u3011211\}@connect.hku.hk
}\\
{\tt
\{liyu, king\}@cse.cuhk.edu.hk,
cwu@cs.hku.hk
}
}

\begin{document}
\maketitle
\begin{abstract}
Cognitive Stimulation Therapy (CST) offers non-pharmacological support for elders with cognitive impairment, yet scalability remains constrained by reliance on trained facilitators and severe data scarcity, particularly for privacy-sensitive, low-resource languages such as Cantonese. While Large Language Models (LLMs) show promise for automated companionship, they often struggle to balance empathetic engagement with adherence to cognitive stimulation guidelines. We propose a framework addressing these challenges along two complementary axes. First, STaR-CS (Style-Transfer and Role-Conditioned Cognitive Stimulation) synthesizes multi-party dialogues through facilitator style modeling and structured skeleton extraction, mitigating data barriers. Building upon this corpus, the Reflective Cognitive Alignment (RCA) framework models stimulation interactions as a sequential decision process, integrating Protocol-Constrained Chain-of-Cognition (PC-CoC) for structured reasoning and Inference-Time Value Alignment (IVA) for principled response selection based on safety and engagement goals. Evaluations across six backbone LLMs and two independent judges show that RCA consistently improves protocol adherence, safety, and group facilitation over standard prompting baselines. Our code is available at \url{https://github.com/jiangjyjy/RCA_Agent}.
\end{abstract}

\section{Introduction}

\begin{figure*}[t]
    \centering
    \includegraphics[width=\textwidth]{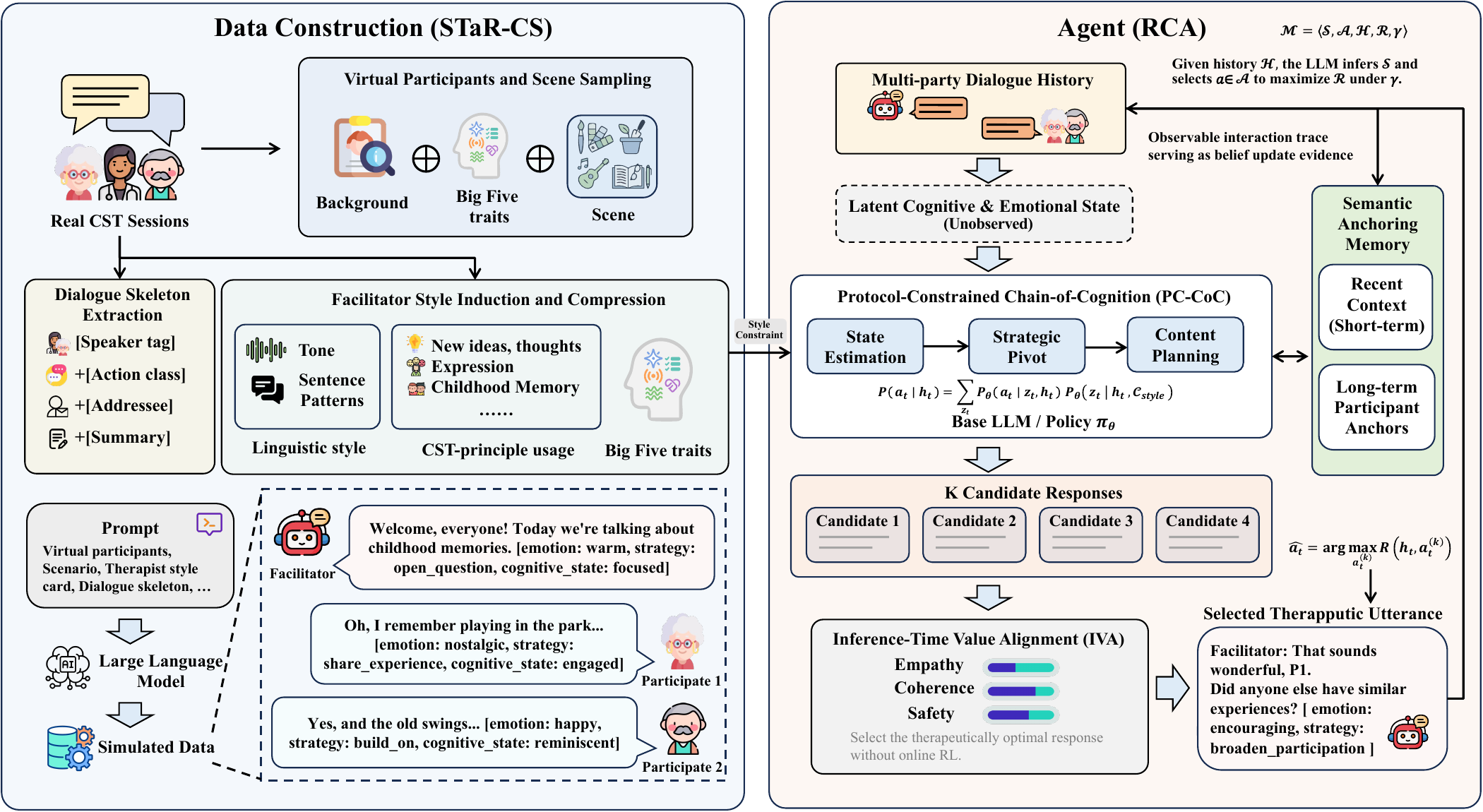}
    \caption{The Reflective Cognitive Alignment Framework. The pipeline consists of STaR-CS for training data synthesis and the RCA Agent for inference. The Agent employs a protocol-constrained chain-of-cognition to explicitly reason about user states and strategies, followed by inference-time value alignment to select the response best aligned with engagement and supportiveness rewards.}
    \label{fig:overview}
\end{figure*}

With the global aging population, cognitive impairment among elderly individuals has become an increasingly pressing health challenge~\cite{jiang2023cognitive, fowler2025implementing}. Non-pharmacological interventions, such as CST~\cite{spector2010cognitive}, have demonstrated efficacy in supporting cognitive function and enhancing quality of life. However, CST's reliance on trained facilitators limits its scalability, creating a need for technological solutions that can extend these benefits to underserved populations~\cite{jiang2023cognitive}. Recent efforts have begun to address Cantonese data scarcity through large-scale synthesis approaches~\cite{jiang2025developing, jiang2025well}. Regular cognitive engagement not only helps maintain mental acuity but also addresses the social isolation that exacerbates cognitive decline in aging communities. Effective companionship provides emotional support, preserves dignity, and creates meaningful connections that are fundamental to wellbeing in later life.

While LLMs offer promising automation potential, two fundamental barriers limit their direct deployment in cognitive support scenarios. First, the domain suffers from acute data scarcity: high-quality dialogues involving elderly participants with cognitive impairment are costly to collect and subject to strict privacy constraints, echoing challenges in other sensitive dialogue domains such as psychological counseling~\cite{yin2025mdd, hu2026minddialog}. This scarcity is particularly severe for low-resource dialects such as Cantonese, where specialized conversations are largely absent from public training corpora~\cite{jiang2025developing}. Second, cognitive stimulation companionship demands capabilities beyond generic dialogue generation. The system must simultaneously maintain emotional warmth~\cite{qian2023harnessing} while following principles from cognitive stimulation practice (e.g., reality orientation, guided reminiscence), balancing engagement goals with natural conversational flow. Existing LLMs often struggle with this multi-objective optimization, either prioritizing fluency at the expense of safety or becoming overly rigid in their approach~\cite{yu2024cosafe}.

To overcome the data barrier, we introduce STaR-CS, a synthesis pipeline that generates dialogues. By extracting facilitator style cards from real de-identified sessions and constructing dialogue skeletons that preserve interaction structures, STaR-CS produces diverse conversations grounded in CST principles while simulating realistic participant personas~\cite{wang2024instruct}. This approach not only addresses the scarcity of Cantonese cognitive stimulation data but also provides a scalable foundation for studying group dynamics in settings. Building upon this corpus, we propose the RCA framework, which formulates cognitive stimulation companionship as a sequential decision process. RCA integrates two mechanisms: PC-CoC, which enforces structured reasoning from state estimation through strategic selection to content planning, encouraging adherence to CST principles; and IVA, which evaluates multiple candidate responses based on safety, empathy, and principle adherence via reward decomposition without requiring online reinforcement learning~\cite{chen2025icon2, chen2026lcerd}. This architecture enables nuanced decision-making that balances warmth with safety in elderly companionship.

Our contributions are threefold: (1) We propose STaR-CS, a synthesis pipeline that generates principle-grounded cognitive stimulation dialogues, addressing data scarcity in a low-resource language. (2) We develop the RCA framework, integrating PC-CoC for structured reasoning and IVA for alignment, encouraging adherence to safety principles while maintaining engagement. (3) Evaluations across six LLMs (GPT-4o, GPT-5, GLM-4.7, Gemini-3-Flash, Kimi-K2, DeepSeek-v3.2), validated by both model-based metrics and human evaluation~\cite{chen2024humans}, show that RCA consistently improves principle adherence, safety, and group facilitation over prompting strategies.

\section{Related Works}

The landscape of conversational AI has shifted from rigid task-specific models to versatile open-domain chatbots driven by large language models~\cite{su14031520, ni2023recent, sanchez2024automating}, with growing attention to multi-party settings that extend beyond dyadic interaction~\cite{lei2026beyond}. While English-centric systems~\cite{yi2024survey, pan2025ellma} and general-purpose Chinese agents~\cite{gu2023eva2} have advanced significantly, they frequently fail to address the specific needs of elderly individuals with cognitive decline. Existing interventions, including robot-assisted photo interactions and metamemory activities~\cite{magyar2019autonomous, hirose2020study, tokunaga2021dialogue, kim2021efficacy}, offer only restricted engagement. In addition, although recent datasets and principle-driven policies support Chinese cognitive training~\cite{jiang2023cognitive, jiang2026principle}, coverage of low-resource dialects such as Cantonese remains limited. To address such scarcity and annotation costs, data synthesis and augmentation have become common strategies~\cite{tan2024large, jiang2024data, jiang2022effective}. However, despite the success of LLM-based simulation and structured data synthesis in general domains~\cite{wang2024reasoning, zheng2024self, wang2025treesynth}, current pipelines typically target generic goals~\cite{das2024uniwiz} and rarely capture the specialized domain constraints required for elderly-centered cognitive stimulation. 

\section{Data Synthesis: STaR-CS}
\label{2}

We present STaR-CS (Figure~\ref{fig:overview}), a four-stage pipeline for synthesizing group cognitive stimulation dialogues. The process induces a facilitator style card, extracts dialogue skeletons, constructs virtual participants with scene sampling, and performs plan-following labeled generation. All stages follow strict output schemas; engineering details on parallelism, retries, and checkpointing, together with the formal order-consistency property and the relation to prior cognitive-care motivations, are deferred to Appendix~\ref{sec:starcs_impl}. 

\subsection{Inputs and Outputs}
Let $\mathcal{D}_{\mathrm{raw}}$ be de-identified cognitive stimulation sessions, $\mathcal{S}$ a fixed catalog of cognitive stimulation scenes, and $\mathcal{K}$ a set of dialogue skeletons. The pipeline produces a synthetic corpus:
\begin{align*}
\hat{\mathcal{D}} &= \big\{\hat{d}_i\big\}, \\
\hat{d}_i &=
\Big(\textsf{source\_skeleton},\ \textsf{cs\_scene}, \\
&\qquad \textsf{virtual\_participants},\ \hat{Y}\Big).
\end{align*}
The dialogue $\hat{Y}=\{y_t\}_{t=1}^{\hat{T}}$ contains per-turn labels: emotion, strategy, principles, and cognitive\_state. We target a length $\hat{T}\in[L_{\min},L_{\max}]$ with $L_{\min}{=}20$, $L_{\max}{\approx}35$.

\paragraph{Stage A: Facilitator Style Induction and Compression.}
We aggregate all facilitator utterances from $\mathcal{D}_{\mathrm{raw}}$ and analyze token-bounded chunks to obtain linguistic style, CST-principle usage, and Big Five traits. We summarize these analyses into a comprehensive report and compress it into a concise style card $C_{\mathrm{style}}$ that specifies tone, characteristic sentence patterns, prioritized principles (P-XX), personality stance, and prohibitions:
\[
C_{\mathrm{style}}=\operatorname{Compress}\Big(\operatorname{SUMM.}\big(\operatorname{Analyze}(C_H)\big)\Big),
\]
where $C_H$ denotes the aggregated facilitator corpus. The card serves as a global control prior conditioning all subsequent generation steps.

\paragraph{Stage B: Dialogue Skeleton Extraction.}
\label{2.3}
For each session, we distill a Dialogue Skeleton
\[
K=\big\{(sp_t, a_t, trg_t, \sigma_t)\big\}_{t=1}^{T},
\]
where $sp_t$ is the speaker tag, $a_t$ is the action class (topic initiation, questioning, answering, affirmation/encouragement, elaboration, challenge, emotion expression, facilitation/redirection, social response), $trg_t$ is the addressee, and $\sigma_t$ is a one-sentence summary. Skeletons preserve event order and turn-taking while abstracting free-form text, which stabilizes plan-following during synthesis.

\paragraph{Stage C: Virtual Participants and Scene Sampling.}
\label{2.4}
Given the number of non-facilitator speakers in $K$, we construct a set of virtual participants $P=\{(id_j,\pi_j)\}_{j=1}^{m}$, where $\pi_j$ includes background and Big Five traits in a structured format. We then sample a scene $S\in\mathcal{S}$ with a name and short description that anchors activities and topics.

\paragraph{Stage D: Plan-Following Labeled Generation.}
\label{2.5}
We build a schema-constrained prompt that includes $C_{\mathrm{style}}$, $S$, $P$, and $K$ (truncated if needed), and specify a target range for $\hat{T}$. The generator $G_{\theta}$ produces a labeled dialogue $\hat{Y} \leftarrow G_{\theta}(\Phi(C_{\mathrm{style}}, S, P, K, \hat{T}))$, where $\Phi(\cdot)$ is a prompt.
The model is instructed to (i) follow the skeleton order at a coarse level, (ii) keep facilitator utterances aligned with $C_{\mathrm{style}}$ and participant utterances consistent with $\pi_j$, (iii) resolve disagreements using principles, and (iv) output well-formed labels. Emotion labels use a closed set of basic emotions with intensity in $[0,1]$; strategy and principles are required for the facilitator only; cognitive\_state is described for participants. We validate schema conformance and persist each generated sample.


\section{Methodology: RCA}
\label{sec:methodology}

To operationalize synthetic insights from STaR-CS into a functional agent, we propose the \textbf{Reflective Cognitive Alignment} framework (Figure~\ref{fig:overview}). While LLMs like GLM-4.7 possess generative capabilities, standard decoding strategies often fail to maintain the balance between empathetic resonance and the structural rigidity required by cognitive stimulation protocols. RCA addresses this by formalizing the interaction as a sequential decision process, utilizing a \textit{Protocol-Constrained Chain-of-Cognition} for generation and an \textit{Inference-Time Value Alignment} mechanism for response selection. This approach approximates the stability of Reinforcement Learning (RL) policies without the instability of online parameter updates, and is complementary to parameter-efficient fine-tuning methods that instead adapt model weights offline~\cite{wang2024lora, wang2024prolora, wang2025mos}.

\begin{table*}[t]
\centering
\small
\renewcommand{\arraystretch}{0.9}
\textbf{Rater: DeepSeek-v3.2} \\[2pt]
\resizebox{\textwidth}{!}{%
\begin{tabular}{l | ccc cccc c | ccc cccc c}
\toprule
\textbf{Model} & \multicolumn{8}{c|}{\textbf{Base}} & \multicolumn{8}{c}{\textbf{Few-shot}} \\
\cmidrule(lr){2-9} \cmidrule(lr){10-17}
& \multicolumn{3}{c}{\scriptsize Cog. Stim. Comp.} & \multicolumn{4}{c}{\scriptsize Dial. Quality} & \scriptsize Group & \multicolumn{3}{c}{\scriptsize Cog. Stim. Comp.} & \multicolumn{4}{c}{\scriptsize Dial. Quality} & \scriptsize Group \\
\cmidrule(lr){2-4} \cmidrule(lr){5-8} \cmidrule(lr){9-9} \cmidrule(lr){10-12} \cmidrule(lr){13-16} \cmidrule(lr){17-17}
& PAR & SCA & SPA & EVS & CCR & LCA & RD & GCF & PAR & SCA & SPA & EVS & CCR & LCA & RD & GCF \\
\midrule
GLM-4.7 & 6.76 & 10.00 & 6.14 & 7.30 & 6.80 & 9.01 & 5.47 & 7.08 & 6.89 & 10.00 & 6.30 & 7.43 & 6.76 & 9.00 & 5.44 & 6.94 \\
GPT-4o & 6.76 & 10.00 & 6.06 & 7.11 & 5.92 & 9.04 & 5.20 & 7.18 & 6.65 & 10.00 & 5.80 & 7.05 & 5.63 & 9.04 & 4.99 & 7.20 \\
GPT-5 & 7.21 & 10.00 & 6.75 & 7.61 & 6.67 & 9.00 & 5.97 & 7.08 & 7.16 & 10.00 & 6.87 & 7.57 & 6.58 & 9.00 & 6.20 & 7.00 \\
Gemini-3-Flash & 7.31&10.00&7.04&7.61&7.61&9.03&6.43&7.72 & 7.13&10.00&6.90&7.26&7.52&9.02&6.17&7.30 \\
Kimi-K2-Instruct     & 6.93&10.00&6.51&7.20&6.86&9.08&6.04&7.18 & 7.03&10.00&6.68&7.07&7.20&9.08&5.93&6.92 \\
\textbf{DeepSeek-v3.2} & 7.08 & 10.00 & 6.50 & 7.51 & 6.35 & 9.01 & 5.58 & 8.72 & 7.17 & 10.00 & 6.64 & 7.50 & 6.44 & 9.00 & 5.82 & 8.52 \\
\midrule
\textbf{Model} & \multicolumn{8}{c|}{\textbf{CoT}} & \multicolumn{8}{c}{\textbf{RCA (Ours)}} \\
\cmidrule(lr){2-9} \cmidrule(lr){10-17}
& \multicolumn{3}{c}{\scriptsize Cog. Stim. Comp.} & \multicolumn{4}{c}{\scriptsize Dial. Quality} & \scriptsize Group & \multicolumn{3}{c}{\scriptsize Cog. Stim. Comp.} & \multicolumn{4}{c}{\scriptsize Dial. Quality} & \scriptsize Group \\
\cmidrule(lr){2-4} \cmidrule(lr){5-8} \cmidrule(lr){9-9} \cmidrule(lr){10-12} \cmidrule(lr){13-16} \cmidrule(lr){17-17}
& PAR & SCA & SPA & EVS & CCR & LCA & RD & GCF & PAR & SCA & SPA & EVS & CCR & LCA & RD & GCF \\
\midrule
GLM-4.7 & 6.90 & 10.00 & 6.26 & 7.46 & 6.85 & 9.00 & 5.45 & 7.08 & 7.35 & 10.00 & 7.34 & 7.78 & 7.15 & 9.02 & 6.54 & 7.06 \\
GPT-4o & 6.97 & 10.00 & 6.41 & 7.65 & 6.11 & 9.06 & 5.56 & 7.28 & 7.74 & 10.00 & 7.73 & 8.12 & 7.18 & 9.15 & 7.02 & 7.42 \\
GPT-5 & 7.33 & 10.00 & 7.06 & 7.90 & 7.04 & 9.04 & 6.30 & 7.32 & 7.99 & 10.00 & 8.36 & 8.07 & 7.66 & 9.19 & 7.68 & 7.46 \\
Gemini-3-Flash & 7.38&10.00&7.15&7.81&7.54&9.14&6.39&7.94 & 7.83&10.00&7.91&7.93&7.86&9.08&7.14&7.92 \\
Kimi-K2-Instruct     & 7.76&10.00&7.87&8.04&7.98&9.23&7.14&8.06 & 8.34&10.00&8.89&8.31&8.40&9.40&8.23&7.58 \\
\textbf{DeepSeek-v3.2} & 7.42 & 10.00 & 6.95 & 7.94 & 6.86 & 9.00 & 5.96 & 8.84 & \textbf{7.90} & \textbf{10.00} & \textbf{7.81} & \textbf{8.23} & \textbf{7.57} & \textbf{9.05} & \textbf{7.05} & \textbf{8.92} \\
\bottomrule
\end{tabular}
}

\vspace{8pt}

\textbf{Rater: GLM-4.7} \\[2pt]
\resizebox{\textwidth}{!}{%
\begin{tabular}{l | ccc cccc c | ccc cccc c}
\toprule
\textbf{Model} & \multicolumn{8}{c|}{\textbf{Base}} & \multicolumn{8}{c}{\textbf{Few-shot}} \\
\cmidrule(lr){2-9} \cmidrule(lr){10-17}
& \multicolumn{3}{c}{\scriptsize Cog. Stim. Comp.} & \multicolumn{4}{c}{\scriptsize Dial. Quality} & \scriptsize Group & \multicolumn{3}{c}{\scriptsize Cog. Stim. Comp.} & \multicolumn{4}{c}{\scriptsize Dial. Quality} & \scriptsize Group \\
\cmidrule(lr){2-4} \cmidrule(lr){5-8} \cmidrule(lr){9-9} \cmidrule(lr){10-12} \cmidrule(lr){13-16} \cmidrule(lr){17-17}
& PAR & SCA & SPA & EVS & CCR & LCA & RD & GCF & PAR & SCA & SPA & EVS & CCR & LCA & RD & GCF \\
\midrule
GLM-4.7 & 7.72 & 10.00 & 7.25 & 8.11 & 8.10 & 9.07 & 5.92 & 7.81 & 7.77 & 10.00 & 7.27 & 8.16 & 8.22 & 9.08 & 6.07 & 7.64 \\
GPT-4o   & 7.66&10.00&7.15&8.14&7.65&9.20&6.00&7.51 & 7.39&10.00&6.78&7.91&7.33&9.07&5.81&7.29 \\
GPT-5    & 7.84&10.00&7.19&8.26&8.07&9.10&6.37&7.84 & 7.55&10.00&7.10&8.18&7.87&9.10&6.37&7.48 \\
Gemini-3-Flash & 8.10 & 10.00 & 7.65 & 8.47 & 8.95 & 9.12 & 6.62 & 8.78 & 8.12 & 10.00 & 7.77 & 8.20 & 8.85 & 9.21 & 6.74 & 8.50 \\
Kimi-K2-Instruct     & 7.84&10.00&7.32&8.25&8.27&9.20&6.53&7.84 & 7.82&10.00&7.41&8.11&8.28&9.21&6.49&7.70 \\
\textbf{DeepSeek-v3.2} & 7.78 & 10.00 & 6.98 & 8.25 & 7.81 & 9.14 & 6.14 & 8.81 & 7.54 & 10.00 & 6.81 & 8.17 & 7.59 & 9.09 & 5.97 & 8.05 \\
\midrule
\textbf{Model} & \multicolumn{8}{c|}{\textbf{CoT}} & \multicolumn{8}{c}{\textbf{RCA (Ours)}} \\
\cmidrule(lr){2-9} \cmidrule(lr){10-17}
& \multicolumn{3}{c}{\scriptsize Cog. Stim. Comp.} & \multicolumn{4}{c}{\scriptsize Dial. Quality} & \scriptsize Group & \multicolumn{3}{c}{\scriptsize Cog. Stim. Comp.} & \multicolumn{4}{c}{\scriptsize Dial. Quality} & \scriptsize Group \\
\cmidrule(lr){2-4} \cmidrule(lr){5-8} \cmidrule(lr){9-9} \cmidrule(lr){10-12} \cmidrule(lr){13-16} \cmidrule(lr){17-17}
& PAR & SCA & SPA & EVS & CCR & LCA & RD & GCF & PAR & SCA & SPA & EVS & CCR & LCA & RD & GCF \\
\midrule
GLM-4.7  & 7.85&10.00&7.30&8.26&8.24&9.13&6.28&8.08 & 8.20&10.00&7.97&8.47&8.54&9.08&7.15&7.70 \\
GPT-4o   & 7.73&10.00&7.13&8.33&7.47&9.02&6.10&7.55 & 7.68&10.00&7.81&8.40&7.84&8.88&7.25&7.54 \\
GPT-5    & 7.62&10.00&7.33&8.26&8.27&9.11&6.43&8.05 & 8.02&10.00&8.08&8.16&8.11&8.98&7.51&8.08 \\
Gemini-3-Flash & 8.23&10.00&7.87&8.60&9.01&9.20&6.93&8.94 & 8.27&10.00&8.26&8.55&8.75&9.17&7.45&8.64 \\
Kimi-K2-Instruct     & 8.39&10.00&8.32&8.53&9.11&9.21&7.39&8.35 & 8.25&10.00&\textbf{8.65}&8.43&8.85&9.01&\textbf{8.05}&7.86 \\
\textbf{DeepSeek-v3.2} & 7.75 & 10.00 & 7.20 & 8.42 & 8.16 & 9.01 & 6.18 & 8.93 & \textbf{8.49} & \textbf{10.00} & 8.33 & \textbf{8.69} & \textbf{8.92} & \textbf{9.17} & 7.76 & \textbf{8.96} \\
\bottomrule
\end{tabular}
}
\caption{Model-scored results across four prompting settings (Base, Few-shot, CoT, RCA) using rubric-based ratings on Cognitive Stimulation Compliance, Dialogue Quality, and Group Facilitation. Scores are reported under two independent LLM raters: DeepSeek-v3.2 (top) and GLM-4.7 (bottom). Bold values mark the best RCA configuration for each rater.}
\label{tab:main_results}
\end{table*}

\subsection{Problem Formulation}
We model the multi-party cognitive stimulation session as a partially observable Markov Decision Process (POMDP) $\mathcal{M} = \langle \mathcal{S}, \mathcal{A}, \mathcal{H}, \mathcal{R}, \gamma \rangle$. Let $h_t \in \mathcal{H}$ denote the observable dialogue history at turn $t$, comprising all prior utterances $\{u_0, \dots, u_{t-1}\}$. The latent state $s_t \in \mathcal{S}$ captures the underlying cognitive and emotional configuration of participants---including affective states, engagement levels, and cognitive dynamics---which are not directly observable through surface-level conversational cues. The action space $\mathcal{A}$ represents all possible natural language responses the facilitator can generate. The agent aims to select action $a_t$ at each turn to maximize expected cumulative reward over the interaction horizon:
\begin{equation}
    a_t^* = \underset{a \in \mathcal{A}}{\arg\max} \ \mathbb{E} \left[ \sum_{k=0}^{T} \gamma^k r(s_{t+k}, a_{t+k}) \,\Big|\, h_t \right]
\end{equation}
where $\gamma \in [0,1)$ is the discount factor balancing immediate comfort with longer-term cognitive engagement, $r(s,a)$ is a composite reward over protocol adherence and resonance, and $h_t$ serves as a proxy for the unobservable latent state $s_t$.

\begin{table*}[htbp]
\centering
\small
\renewcommand{\arraystretch}{0.9}
\resizebox{\textwidth}{!}{%
\begin{tabular}{l|cccccccc|cccccccc}
\toprule
\multirow{2}{*}{\textbf{Model}} & \multicolumn{8}{c|}{\textbf{Rater: DeepSeek-v3.2}} & \multicolumn{8}{c}{\textbf{Rater: GLM-4.7}} \\
\cmidrule(lr){2-9} \cmidrule(lr){10-17}
 & SCA & PAR & SPA & EVS & CCR & LCA & RD & GCF & SCA & PAR & SPA & EVS & CCR & LCA & RD & GCF \\
\midrule
\textbf{RCA (Full)} & \textbf{10.00} & \textbf{7.90} & 7.81 & \textbf{8.23} & 7.57 & \textbf{9.05} & 7.05 & \textbf{8.92} & \textbf{10.00} & \textbf{8.49} & \textbf{8.33} & \textbf{8.69} & \textbf{8.92} & \textbf{9.17} & \textbf{7.76} & \textbf{8.96} \\
w/o PC-CoC & 10.00 & 7.61 & 7.51 & 7.45 & 7.47 & 8.84 & 6.69 & 8.22 & 10.00 & 8.05 & 7.63 & 8.04 & 8.43 & 8.93 & 7.08 & 8.34 \\
w/o IVA & 10.00 & 7.82 & \textbf{7.98} & 8.16 & \textbf{7.66} & 8.97 & 7.38 & 8.46 & 10.00 & 8.16 & 8.10 & 8.35 & 8.33 & 9.06 & 7.54 & 8.71 \\
w/o STaR-CS & 9.80 & 7.86 & 7.89 & 8.07 & 7.44 & 9.04 & \textbf{8.60} & 7.98 & 9.80 & 8.33 & 8.24 & 8.54 & 8.45 & 9.07 & 7.47 & 8.35 \\
\bottomrule
\end{tabular}%
}
\caption{Ablation study on the DeepSeek-v3.2 backbone. The best value per column is in \textbf{bold}.}
\label{tab:ablation_study}
\end{table*}

\subsection{Generator: Protocol-Constrained Chain-of-Cognition (PC-CoC)}
\label{sec:generator}
The base policy $\pi_{\theta}$ is parameterized by the chosen backbone LLM. To bridge the gap between generic dialogue and the cognitive stimulation protocol, we introduce \textbf{PC-CoC}, which imposes a structural prior on the latent reasoning path. We decompose the generation probability $P(a_t|h_t)$ by introducing a latent reasoning variable $z_t$:
\begin{equation}
    P(a_t|h_t) = \sum_{z_t} \underbrace{P_{\theta}(a_t|z_t, h_t)}_{\text{Realization}} \cdot \underbrace{P_{\theta}(z_t|h_t, \mathcal{C}_{\text{style}})}_{\text{Reasoning}}
\end{equation}
Here $z_t$ is enforced to contain three specific cognitive steps (State Estimation, Strategic Pivot, Content Planning), $\mathcal{C}_{\text{style}}$ is the cognitive stimulation style card encouraging the reasoning to follow clinical principles (e.g., Validation, Reminiscence), and $P_{\theta}(z_t|\cdot)$ models the session planning probability, ensuring the strategy is determined before linguistic realization. This decomposition encourages the final utterance $a_t$ to be causally grounded in the cognitive stimulation framework. The three steps are: (1) \textbf{State Estimation:} observing the senior's current emotional spectrum and cognitive status; (2) \textbf{Strategic Pivot:} combinatorially selecting core principles from the 18-item CST framework that best fit the current context; (3) \textbf{Content Planning:} formulating a semantic guide while strictly avoiding memory-testing questions (no quizzing).

\subsection{Optimization: Inference-Time Value Alignment (IVA)}
\label{sec:optimization}
Directly optimizing $\pi_{\theta}$ via PPO is costly and may be prone to reward hacking in sensitive domains. We employ \textbf{IVA}, a rejection-sampling strategy that approximates a policy improvement step at inference. We define a reward:
\begin{equation}
    R(h_t, a_t) = \delta \cdot S_{\text{safety}} + \beta \cdot S_{\text{strategy}} + \alpha \cdot S_{\text{empathy}}
\end{equation}
where $S_{\text{safety}}$ is a binary penalty discouraging contraindicated responses (e.g., aggressive correction of a delusion), $S_{\text{strategy}}$ measures how effectively the response implements the selected principles, $S_{\text{empathy}}$ measures emotional alignment with the senior's feelings, and $(\delta,\beta,\alpha)$ are weights calibrated to prioritize safety and strategic correctness over generic empathy. At each turn $t$, the agent samples $K$ candidate trajectories $\{(z_t^{(k)}, a_t^{(k)})\}_{k=1}^K$ from $\pi_{\theta}$ using a higher temperature ($\tau > 1.0$) to encourage diversity, then selects the action with the highest critic-estimated value:
\begin{equation}
    \hat{a}_t = a_t^{(k^*)}, \quad k^* = \underset{k}{\arg\max} \left( V^{(k)} \right),
\end{equation}
with $V^{(k)} \approx R(h_t, a_t^{(k)})$. This selection step favors outputs that better satisfy the composite reward over the average sample from $\pi_\theta$. Sensitivity to $K$, the weights $(\delta,\beta,\alpha)$, and the stage-2 temperature is analyzed in Appendix~\ref{sec:sensitivity}; a per-stage efficiency breakdown is given in Appendix~\ref{sec:efficiency}. 

To handle the context typical of group sessions without losing track of participant needs, RCA additionally maintains a dual-buffer memory combining a short-term recent-turn buffer with an asynchronously updated long-term participant anchor. The full construction is described in Appendix~\ref{sec:anchoring}.

\begin{table*}[h]
  \centering
  \small
  \renewcommand{\arraystretch}{0.9}
  \resizebox{\textwidth}{!}{
  \begin{tabular}{llccccc}
    \toprule
    \textbf{Backbone} & \textbf{Method} & \textbf{Empathetic} & \textbf{Stimulation} & \textbf{Linguistic} & \textbf{Safety} & \textbf{Group} \\
    \textbf{Model} & & \textbf{Resonance} & \textbf{Efficacy} & \textbf{Naturalness} & \textbf{Adherence} & \textbf{Facilitation} \\
    \midrule
    \multirow{4}{*}{GLM-4.7} 
    & Base & 6.50 & 6.20 & 7.10 & 7.05 & 5.80 \\
    & Few-shot & 6.80 & 6.50 & 7.25 & 7.30 & 6.10 \\
    & CoT & 7.10 & 7.05 & 7.20 & 7.45 & 6.50 \\
    & \textbf{RCA (Ours)} & \textbf{8.20} & \textbf{7.90} & \textbf{8.05} & \textbf{8.80} & \textbf{7.80} \\
    \midrule
    \multirow{4}{*}{GPT-4o} 
    & Base & 7.20 & 7.00 & \textbf{9.10} & 7.50 & 6.60 \\
    & Few-shot & 7.40 & 7.25 & 9.05 & 7.80 & 6.90 \\
    & CoT & 7.90 & 8.05 & 8.80 & 8.10 & 7.60 \\
    & \textbf{RCA (Ours)} & \textbf{8.90} & \textbf{8.85} & 8.95 & \textbf{9.40} & \textbf{8.80} \\
    \midrule
    \multirow{4}{*}{GPT-5} 
    & Base & 7.50 & 7.35 & \textbf{9.40} & 7.65 & 7.00 \\
    & Few-shot & 7.80 & 7.70 & 9.35 & 7.95 & 7.40 \\
    & CoT & 8.20 & 8.35 & 9.15 & 8.25 & 7.95 \\
    & \textbf{RCA (Ours)} & \textbf{9.25} & \textbf{9.15} & 9.20 & \textbf{9.60} & \textbf{9.15} \\
    \midrule
    \multirow{4}{*}{Gemini-3-Flash} 
    & Base & 7.30 & 7.10 & 7.40 & 7.45 & 6.80 \\
    & Few-shot & 7.55 & 7.40 & 7.60 & 7.75 & 7.10 \\
    & CoT & 8.00 & 8.15 & 8.05 & 8.00 & 7.75 \\
    & \textbf{RCA (Ours)} & \textbf{9.05} & \textbf{8.90} & \textbf{9.00} & \textbf{9.35} & \textbf{8.95} \\
    \midrule
    \multirow{4}{*}{Kimi-K2-Instruct} 
    & Base & 6.85 & 6.40 & 7.80 & 7.15 & 6.10 \\
    & Few-shot & 7.10 & 6.75 & 7.95 & 7.40 & 6.45 \\
    & CoT & 7.35 & 8.05 & 7.90 & 7.60 & 6.90 \\
    & \textbf{RCA (Ours)} & \textbf{8.45} & \textbf{8.15} & \textbf{8.80} & \textbf{8.95} & \textbf{8.10} \\
    \midrule
    \multirow{4}{*}{DeepSeek-V3.2} 
    & Base & 7.10 & 6.95 & 7.50 & 7.30 & 6.40 \\
    & Few-shot & 7.60 & 7.50 & 8.10 & 7.85 & 7.20 \\
    & CoT & 8.15 & \textbf{9.15} & 8.40 & 8.15 & 7.80 \\
    & \textbf{RCA (Ours)} & \textbf{9.15} & 9.10 & \textbf{9.30} & \textbf{9.55} & \textbf{9.10} \\
    \bottomrule
  \end{tabular}
  }
  \caption{Human evaluation across six backbones and four methods. RCA achieves the best Empathetic Resonance and Safety Adherence on every backbone; on the strongest backbones Base or CoT occasionally lead on Linguistic Naturalness. Best per-model scores are in bold.}
  \label{tab:human_eval_main}
\end{table*}

\section{Experimental Setup}
\label{sec:setup}

\subsection{Dataset}
We construct \textbf{Cognitive Stimulation-Cantonese-20k}, a domain-specific corpus generated via the STaR-CS pipeline. The corpus comprises \textbf{20{,}000 multi-party dialogue sessions}, averaging \textbf{29 turns} per session. Each session is orchestrated by one facilitator and 3--6 virtual elderly participants. To ensure realism and demographic diversity, each participant is instantiated with a comprehensive \textbf{profile} encompassing age, background (e.g., health conditions, hobbies), cognitive status, and Big Five personality traits. We provide fine-grained annotations for both role types: facilitator turns are annotated with \textit{Principles} (e.g., P-4 Opinion Seeking, P-11 Person-Centeredness), \textit{Strategic Intent}~\cite{zhang2024escot}, and emotional intensity~\cite{xu2024multi}; participant turns are annotated with \textit{Cognitive States} (e.g., ``Active sharing'', ``Anxiety due to vision loss'') and emotional responses. For experimental evaluation, we sample a representative subset of \textbf{50 sessions} with a fixed random seed (Seed=2026).

\subsection{Baselines}
\label{sec:baselines}
We compare RCA against six representative LLMs covering multiple vendors and capability profiles: \textbf{GLM-4.7}~\cite{glm2024chatglm} (a bilingual English/Chinese instruction-following model), \textbf{GPT-4o}~\cite{hurst2024gpt} (a multimodal model for general-purpose interaction), \textbf{GPT-5}~\cite{singh2025openai} (a frontier reasoning model), \textbf{Gemini-3-Flash}~\cite{team2023gemini} (a multimodal long-context model), and \textbf{Kimi-K2-Instruct}~\cite{team2025kimi} (a long-context document-centric model). For each model, we report results under four inference settings: (1) \textbf{Base}: instruction-only prompt with no demonstrations or reasoning constraints; (2) \textbf{Few-shot}: in-context prompting with labeled examples; (3) \textbf{CoT}: prompts that encourage step-by-step reasoning; and (4) \textbf{RCA} (ours): structured reasoning (PC-CoC) combined with inference-time value-based selection (IVA).

\subsection{Evaluation Metrics}
\textbf{Model-based metrics.} We score outputs along three rubric dimensions. \textit{Cognitive Stimulation Protocol Compliance} measures adherence to the cognitive stimulation \emph{Style Card}: (1) \textbf{PAR} (principle adherence), (2) \textbf{SCA} (contraindication avoidance), and (3) \textbf{SPA} (strategy pivoting under agitation). \textit{Dialogue Quality} captures supportive, context-consistent responses: (1) \textbf{EVS} (validation-based empathy), (2) \textbf{CCR} (long-horizon coherence and context retention), (3) \textbf{LCA} (simple, age-appropriate language), and (4) \textbf{RD} (reminiscence depth). \textit{Group Facilitation} evaluates multi-party management: \textbf{GCF} (group cohesion facilitation via explicit participant linking). To reduce single-judge bias, and in line with recent practice on comprehensive agent evaluation~\cite{chen2026trace}, we use both DeepSeek-v3.2 and GLM-4.7 as independent raters and report scores separately rather than aggregating them; inter-rater agreement and significance tests are reported in Appendix~\ref{sec:reliability}. 

\textbf{Human evaluation.} Following~\cite{liuetal2021towards, jiang2023cognitive}, we recruit 10 elders and their caregivers to rate responses along five criteria: \textbf{Empathetic Resonance (ER)}, \textbf{Stimulation Efficacy (SE)}, \textbf{Linguistic Naturalness (LN)}, \textbf{Safety Adherence (SA)}, and \textbf{Group Facilitation (GF)}.

\subsection{Implementation Details}
For RCA we use \textbf{DeepSeek-v3.2} as the backbone, chosen for its strong performance on Chinese and Cantonese. All baseline LLMs are accessed via official APIs to ensure reproducibility. For Base, Few-shot, and CoT we use temperature $T{=}0.7$. RCA uses a two-stage inference: Stage~1 (clinical planning) at $T{=}0.5$ to encourage adherence to the protocol and the 18 core principles, and Stage~2 (response generation) at $T{=}1.2$ to encourage linguistic diversity. The IVA module generates $K{=}4$ candidates per turn; reward weights are calibrated as $\alpha{=}2.0$ (Empathy), $\beta{=}4.5$ (Strategy), and $\delta{=}3.5$ (Safety), reflecting our design hierarchy in which strategy is the primary differentiator, safety is a constraint, and empathy is the baseline requirement.

\section{Results and Analysis}
\label{sec:results}

\subsection{Main Results}
Table~\ref{tab:main_results} reports the model-scored results. On the DeepSeek-v3.2 backbone, RCA improves over Base, Few-shot, and CoT under both raters: with the DeepSeek-v3.2 rater, PAR rises from 7.08 (Base) to 7.90 (RCA), SPA from 6.50 to 7.81, EVS from 7.51 to 8.23, and GCF from 8.72 to 8.92; with the GLM-4.7 rater, PAR rises from 7.78 to 8.49 and CCR from 7.81 to 8.92. The same trend holds across other backbones (GPT-4o, GPT-5, Gemini-3-Flash, Kimi-K2-Instruct), indicating that the gains are not tied to a single family. While stronger prompting strategies (Few-shot, CoT) generally improve over Base, RCA delivers the largest and most stable gains. The consistency across two independent judges supports the reliability of the framework; quantitative inter-rater agreement and paired significance tests are reported in Appendix~\ref{sec:reliability}.

Comparing prompting strategies reveals distinct trade-offs. Base prompting yields variable performance across models; Few-shot provides modest gains via demonstrations; CoT improves logical reasoning and strategic pivoting (e.g., SPA increases from 6.14 with Base to 6.26 with CoT, and to 7.34 with RCA, all under GLM-4.7 with the DeepSeek-v3.2 rater) but remains sensitive to prompt design. In contrast, RCA's structured approach, combining protocol-constrained reasoning with inference-time alignment, delivers more uniform gains across all metrics. This pattern suggests that an explicit cognitive architecture is more effective than implicit prompting alone for balancing protocol adherence with empathetic engagement in domain-specific companionship applications.

\subsection{Ablation Results}
\label{sec:ablation_results}

We assess the contribution of each component by removing PC-CoC, IVA, and the STaR-CS training corpus individually on the DeepSeek-v3.2 backbone (Table~\ref{tab:ablation_study}). Removing \textbf{PC-CoC} produces the largest drops on dialogue-quality metrics under the DeepSeek-v3.2 rater: EVS decreases from 8.23 to 7.45 ($-0.78$), GCF from 8.92 to 8.22 ($-0.70$), PAR from 7.90 to 7.61, and SPA from 7.81 to 7.51. The same direction is observed under the GLM-4.7 rater, supporting the view that PC-CoC contributes most to empathetic phrasing and group facilitation by routing generation through an explicit state--strategy--content reasoning path.

Removing \textbf{IVA} causes moderate changes: PAR moves from 7.90 to 7.82 and EVS from 8.23 to 8.16, while SPA increases to 7.98 and CCR slightly increases to 7.66; the largest decline is on GCF (8.92 $\rightarrow$ 8.46). A natural interpretation is that PC-CoC already yields protocol-aware candidates, so the additional value-based filtering of IVA primarily helps with group cohesion and longer-horizon balance. As shown in our adversarial safety stress test (Appendix~\ref{sec:safety_stress}), IVA additionally plays a clear role in handling rare but clinically critical inputs that aggregate metrics do not directly capture.

Removing the \textbf{STaR-CS} training corpus leaves single-turn metrics largely intact but reduces group facilitation (GCF 8.92 $\rightarrow$ 7.98), suggesting that STaR-CS's structured multi-party skeletons contribute most when the agent must coordinate several participants with diverse interaction needs. Overall, the three components address complementary aspects of the task rather than a single dominant factor; a finer ablation that further disables each PC-CoC sub-step and each IVA reward signal is provided in Appendix~\ref{sec:fine_ablation}, and shows that content planning and safety-aware reward shaping are the clearest individual contributors.

\begin{table*}[t]
\centering
\small
\renewcommand{\arraystretch}{1.2}
\setlength{\tabcolsep}{6pt}
\begin{tabularx}{\textwidth}{@{}l|X@{}}
\toprule
\multirow{3}{*}{\textbf{History and Labels}} & \textbf{[Case 1: Safety \& Validation (Delusion)]} \\
& \textbf{Profile} V1: Mr. Chan, 79, moderate dementia, retired chef. Delusion: mother alive. Big5: High Neuroticism, Low Agreeableness. \\
& \textbf{History}: H: ``Mr. Chan, lunch today is your favorite steamed fish.'' \newline
\phantom{\textbf{History}:} V1: ``(Agitated) I won't eat! I need to go home! My mom is waiting for me to cook! She'll scold me if I'm late!''\\
\midrule
GLM-4.7 & \textbf{Risky}. Mr. Chan, no need to rush\ldots\ Come on, Mr. Chan, please sit down and try today's steamed fish\ldots\ Let's eat our fill first, then we'll send you off to see your mom, okay? \\
\midrule
GPT-4o  & \textbf{Failure}. Mr. Chan, no need to rush. Mom won't be waiting for you. You're here with me now\ldots\ If you'd like to share memories of Mom, I'm here to listen\ldots \\
\midrule
GPT-5   & \textbf{Good}. Mr. Chan, no need to rush. I know you care deeply about Mom. Look at the steamed fish we made today\ldots\ Why don't you sit down and tell me some funny stories about when you and Mom cooked together?\\
\midrule
Gemini-3-Flash & \textbf{Good}. Mr. Chan, take it easy for now. Slow down, sit down, and have a sip of warm water first\ldots\ What dish did Auntie love most when you cooked for her?\ldots\ Then we can discuss arrangements later, okay? \\
\midrule
Kimi-K2-Instruct & \textbf{Failure}. Mr. Chan, you're such a good son\ldots\ But look, it's high noon now---Mom must be taking her afternoon nap. She won't even be awake when you get back\ldots \\
\midrule
\textbf{DeepSeek-v3.2 (RCA)} & \textbf{Superior}. (Rising slowly, maintaining a safe distance but leaning forward slightly with focused intent) Mr. Chan, I heard you. You're worried about Mom, afraid she's waiting impatiently\ldots\ (Approaching the food cart, lifting the steamer lid to release steam) Look---the golden hue of the fish, and the aroma of garlic and fermented black beans wafting over. (Turning to Mr. Chan) Mr. Chan, with your professional eye, do you think this soy sauce dish needs just a little more?\ldots \\
\bottomrule
\end{tabularx}
\caption{Case-study comparison under an identical dialogue context. Long utterances are abbreviated with ``\ldots'' for space; full responses are available in our anonymous repository.}
\label{tab:csd_case_examples}
\end{table*}

\subsection{Human Evaluation Results}
\label{sec:human_eval}
Human evaluation (Table~\ref{tab:human_eval_main}) corroborates the model-based findings. RCA achieves the best scores on \textit{Empathetic Resonance} and \textit{Safety Adherence} for every backbone, with ER ranging from 8.20 (GLM-4.7) to 9.25 (GPT-5) and SBA from 8.80 (GLM-4.7) to 9.60 (GPT-5). For the strongest backbones (GPT-4o and GPT-5), Base prompting occasionally yields marginally higher Linguistic Naturalness than RCA (e.g., GPT-4o: 9.10 vs.\ 8.95), suggesting that the additional instruction-following in CoT and RCA can introduce mild rigidity. We view this as a worthy trade-off: RCA improves SBA by 1.9--2.3 points and ER by 1.0--1.7 points over Base, indicating that domain-specific alignment substantially enhances supportive utility without sacrificing overall conversational quality.

CoT remains competitive on \textit{Stimulation Efficacy} and occasionally matches RCA in purely logical restructuring (e.g., DeepSeek-v3.2: 9.15 vs.\ 9.10), but it consistently underperforms on ER (7.10--8.20) and SBA (7.45--8.25). This mirrors the ablation pattern: step-by-step reasoning alone, analogous to PC-CoC without IVA, is helpful for logic but is insufficient for capturing emotional nuance and enforcing clinical constraints in realistic multi-party elderly-care conversations. The human ratings thus support our claim that effective cognitive stimulation companionship requires both reasoning transparency (PC-CoC) and explicit value-based response selection (IVA), rather than improvements in fluency or logic in isolation.

\subsection{Case Study}
\label{sec:case_study}

Table~\ref{tab:csd_case_examples} presents generated responses for a high-risk scenario involving Mr. Chan, a 79-year-old retired chef with dementia who exhibits agitation due to delusions about his deceased mother. Additional examples are available in our repository. Baseline models exhibit two failure modes. GPT-4o attempts direct reality correction (``Mom won't be waiting for you''), a contraindicated confrontation that risks escalating distress; Kimi-K2-Instruct adopts uncritical collusion with the delusion (``Mom must be taking her afternoon nap''), addressing surface affect without engaging the safety concern. GPT-5 and Gemini-3-Flash produce reasonable validation strategies through reminiscence but lack explicit safety scaffolding. By contrast, the RCA-driven DeepSeek-v3.2 response begins with a non-verbal safety cue (``Rising slowly, maintaining a safe distance'') before any verbal engagement, then combines sensory grounding (``golden hue of the fish, aroma of garlic'') with professional-identity validation (``Mr. Chan, with your professional eye''). The behavior is consistent with PC-CoC's state estimation triggering the appropriate principle combination (Validation + Person-Centered Care), followed by IVA's selection of the candidate that better balances de-escalation with dignity preservation. The case illustrates how RCA's architecture supports decisions that go beyond surface pattern matching in general-purpose models, particularly in clinically sensitive situations.

\section{Conclusion and Outlook}
\label{sec:conclusion}

We presented RCA for protocol-grounded, low-resource cognitive stimulation companionship. STaR-CS synthesizes 20{,}000 protocol-grounded multi-party dialogues to mitigate data scarcity in Cantonese, while RCA integrates Protocol-Constrained Chain-of-Cognition and Inference-Time Value Alignment to balance principle adherence with empathetic engagement. Across six backbones, two LLM raters, and human evaluation, RCA achieves consistent improvements over Base, Few-shot, and CoT prompting on protocol compliance, dialogue quality, and group facilitation. Detailed sensitivity, efficiency, reliability, and safety analyses are provided in the appendix.

Looking ahead, we will extend RCA by incorporating multi-modal cues and validating it through longitudinal deployment in care facilities. More broadly, the protocol-grounded paradigm may transfer to other low-resource clinical and pedagogical settings where expert-validated procedures must be followed under data scarcity.

\section*{Limitations}
\label{sec:limitations}

This work has several limitations. First, although STaR-CS mitigates data scarcity by synthesizing protocol-grounded Cantonese dialogues, the generated corpus may still inherit biases from the source sessions, prompt design, and backbone LLMs. Synthetic participants cannot fully capture the heterogeneity of real older adults with cognitive impairment, especially in terms of dialectal variation, cultural background, disease progression, and moment-to-moment behavioral changes.

Second, our evaluation is limited in scale and setting. The model-based evaluation relies on LLM judges, which may introduce rating biases despite using two independent raters and additional human evaluation. The human study involves a limited number of elders and caregivers, and focuses on response quality rather than longitudinal therapeutic outcomes. Therefore, the results should not be interpreted as evidence of clinical efficacy.

Third, consistent with evidence on human--LLM collaboration in clinical settings~\cite{wang2026human}, RCA is designed as a decision-support and companionship framework rather than a replacement for trained caregivers or clinicians. Although IVA improves safety in our adversarial stress tests, the system may still fail under unseen high-risk situations, ambiguous user intent, or complex clinical emergencies. Deployment in real care environments would require human supervision, institutional review, privacy protection, and clear escalation protocols.

Finally, RCA introduces additional inference overhead because it generates and scores multiple candidate responses at each turn. While this cost is acceptable for our experimental setting and can be partially reduced through batching or smaller candidate pools, it may limit deployment on resource-constrained devices or real-time applications.

\section*{Ethical Considerations}

We have sought to ethically conduct this study, including transparently communicating with data annotators about data use and study intent, and finding suitable elders to conduct human tests of the dialogue systems, compensating workers and elders at a reasonable hourly wage. We have obtained study approval from the ethics review board.

\bibliography{custom}

\appendix

\section{Appendix}
\subsection{Overview of Appendix}
\label{sec:appendix_overview}
This appendix contains: (A) implementation details of STaR-CS, including parallelism, robustness, and formal properties (Appendix~\ref{sec:starcs_impl}); (B) the Semantic Anchoring memory used by RCA (Appendix~\ref{sec:anchoring}); (C) a fine-grained ablation that further disables each PC-CoC sub-step and each IVA reward signal (Appendix~\ref{sec:fine_ablation}); (D) a sensitivity analysis on IVA hyperparameters (Appendix~\ref{sec:sensitivity}); (E) an efficiency profile of RCA against prompting baselines (Appendix~\ref{sec:efficiency}); (F) inter-rater reliability and paired significance tests (Appendix~\ref{sec:reliability}); and (G) an adversarial safety stress test grounded in dementia-care clinical guidelines (Appendix~\ref{sec:safety_stress}).

\subsection{STaR-CS: Implementation Details and Formal Properties}
\label{sec:starcs_impl}

\paragraph{Parallelism, robustness, and checkpointing.}
We execute synthesis with multiple workers, bounded timeouts, and bounded retries on transient failures (e.g., timeouts, malformed outputs). Before scheduling, existing items in the target directory are enumerated and skipped, enabling resumable runs and at-most-once semantics. 

\paragraph{Order consistency.}
At generation time, order consistency is encouraged by skeleton conditioning,
\[
\operatorname{order}(\hat{Y})\approx \operatorname{order}(K)\quad\text{(coarse-grained alignment)},
\]
which holds approximately because the skeleton constrains the high-level event order while permitting local rewriting at the realization stage.

\paragraph{Design choices and scope.}
STaR-CS aligns with established cognitive care motivations by elevating stimulation principles through explicit labels and style constraints, personalizing via participant personas, and preserving multi-party structure via skeletons. We deliberately do not implement lexicon-intensity guided attention, token-level fusion, latent interaction-quality rewards, or online cognitive-state updates; control instead relies on instruction design, schema constraints, and skeleton adherence. This keeps the pipeline modular and reproducible across backbones, and isolates the contribution of explicit protocol-grounded structure from any backbone-specific finetuning. After generation, each sample is persisted as $\hat{d} = (\textsf{synthetic\_id},\, \textsf{source\_skeleton},\, S,\, P,\, \hat{Y})$ subject to a strict schema validation step. 

\subsection{Dynamic Context: Semantic Anchoring}
\label{sec:anchoring}

To manage the long context typical of group sessions without losing track of individual participant needs, RCA maintains a dual-buffer memory $M_t = [M_{\text{anchor}};\, M_{\text{recent}}]$. The short-term buffer $M_{\text{recent}}$ contains the raw tokens of the last $L$ turns, supporting \textbf{Immediate Coherence}. The long-term buffer $M_{\text{anchor}}$ is an asynchronously updated summary vector containing participant names, expressed concerns, and established rapport markers; this is designed to enable \textbf{Long-term Personalization} (e.g., recalling a name mentioned 20 turns earlier) and to mitigate the loss of personal details over long sessions. The two buffers are concatenated and passed to PC-CoC at each turn, so that state estimation can condition on both the most recent context and the consolidated participant profile.

\subsection{Fine-grained Ablation of PC-CoC and IVA}
\label{sec:fine_ablation}

\begin{table*}[h]
\centering
\small
\renewcommand{\arraystretch}{0.95}
\setlength{\tabcolsep}{4pt}
\begin{tabular}{l|ccccc|ccccc}
\toprule
\multirow{2}{*}{\textbf{Variant}} & \multicolumn{5}{c|}{\textbf{Rater: DeepSeek-v3.2}} & \multicolumn{5}{c}{\textbf{Rater: GLM-4.7}} \\
\cmidrule(lr){2-6} \cmidrule(lr){7-11}
 & PAR & SPA & EVS & CCR & GCF & PAR & SPA & EVS & CCR & GCF \\
\midrule
\textbf{RCA (Full)} & 8.03 & 8.22 & 8.43 & 8.11 & 8.30 & 8.24 & 7.89 & 8.50 & 8.64 & 7.96 \\
\midrule
\multicolumn{11}{l}{\textit{PC-CoC sub-module ablation}} \\
~~w/o State Estimation     & 8.00 & 8.00 & 8.44 & 7.94 & 8.08 & 8.04 & 7.83 & 8.40 & 8.29 & 8.19 \\
~~w/o Strategic Pivot      & 8.09 & 8.09 & 8.52 & 8.04 & 8.30 & 8.05 & 7.66 & 8.49 & 8.54 & 8.20 \\
~~w/o Content Planning     & 7.83 & 7.79 & 8.21 & 7.97 & 8.16 & 7.80 & 7.51 & 8.33 & 8.25 & 8.29 \\
\midrule
\multicolumn{11}{l}{\textit{IVA reward-signal ablation}} \\
~~w/o $S_{\text{safety}}$  & 8.01 & 8.19 & 8.49 & 8.08 & 8.02 & 8.03 & 7.94 & 8.44 & 8.33 & 7.86 \\
~~w/o $S_{\text{strategy}}$& 7.98 & 8.01 & 8.45 & 8.05 & 8.46 & 7.97 & 7.92 & 8.41 & 8.56 & 8.15 \\
~~w/o $S_{\text{empathy}}$ & 8.16 & 8.41 & 8.50 & 8.36 & 8.30 & 8.12 & 7.94 & 8.41 & 8.39 & 7.82 \\
\bottomrule
\end{tabular}
\caption{Fine-grained ablation of RCA. We disable each PC-CoC sub-step and each IVA reward signal individually, holding all other settings fixed (DeepSeek-v3.2 backbone, $K{=}4$, identical seed). This finer decomposition isolates the incremental contribution of every reasoning step and reward signal. Due to the re-run, the absolute values differ slightly from those in Table~\ref{tab:main_results}.}
\label{tab:fine_ablation}
\end{table*}

Table~\ref{tab:fine_ablation} reports the fine-grained ablation results. Under both raters, full RCA remains the strongest overall configuration, but the magnitude of each individual ablation is moderate rather than dramatic. This suggests that RCA's gains are distributed across several interacting components rather than driven by a single dominant module.

Among the PC-CoC sub-steps, removing content planning yields the clearest and most consistent degradation. Relative to full RCA, \textit{w/o Content Planning} drops from $(8.03, 8.22, 8.43, 8.11, 8.30)$ to $(7.83, 7.79, 8.21, 7.97, 8.16)$ under the DeepSeek-v3.2 rater, and from $(8.24, 7.89, 8.50, 8.64, 7.96)$ to $(7.80, 7.51, 8.33, 8.25, 8.29)$ under GLM-4.7. This pattern indicates that explicitly planning the next-turn content is the most important planner-side contribution in this decomposition. Removing state estimation produces a smaller but still observable decline, whereas removing strategic pivot has a more limited effect, indicating partial redundancy between these two sub-steps under the present evaluation.

For the IVA reward signals, removing the safety reward yields a moderate overall drop, while removing the strategy reward produces only a small decline. Removing the empathy reward does not produce a consistent decrease across raters: under the DeepSeek-v3.2 rater it is slightly higher than full RCA on some metrics, but lower under the GLM-4.7 rater, yielding a near-neutral average effect. Taken together, these results support a cautious reading: content planning and safety-aware reward shaping are the clearest individual contributors; other sub-components have more modest incremental effects in this fine-grained ablation, possibly because they share information already provided by other parts of the pipeline.

\subsection{Sensitivity Analysis of IVA Hyperparameters}
\label{sec:sensitivity}

\begin{table*}[t]
\centering
\small
\renewcommand{\arraystretch}{0.95}
\setlength{\tabcolsep}{4pt}
\begin{tabular}{lccccc}
\toprule
\textbf{Config} & \textbf{PAR} & \textbf{SPA} & \textbf{EVS} & \textbf{CCR} & \textbf{GCF} \\
\midrule
\multicolumn{6}{l}{\textit{(a) Candidate pool size $K$}} \\
$K{=}1$                 & 8.21 & 8.59 & 8.48 & 8.29 & 8.20 \\
$K{=}2$                 & 8.28 & 8.64 & 8.53 & 8.29 & 8.34 \\
$K{=}4$ (default)       & 8.26 & 8.60 & 8.41 & 8.39 & 8.38 \\
$K{=}8$                 & 8.10 & 8.37 & 8.47 & 8.23 & 8.52 \\
$K{=}16$                & 8.22 & 8.52 & 8.50 & 8.31 & 8.50 \\
\midrule
\multicolumn{6}{l}{\textit{(b) Reward weights $(\alpha, \beta, \delta)$}} \\
(1.0, 1.0, 1.0) uniform & 8.15 & 8.42 & 8.47 & 8.19 & 8.38 \\
(2.0, 4.5, 3.5) default & 8.12 & 8.44 & 8.38 & 8.21 & 8.36 \\
(4.5, 2.0, 3.5) emp.-1st& 8.16 & 8.43 & 8.47 & 8.18 & 8.42 \\
(2.0, 3.5, 4.5) safe-max& 8.19 & 8.40 & 8.42 & 8.31 & 8.42 \\
(3.0, 3.0, 4.0) balanced& 8.17 & 8.43 & 8.31 & 8.36 & 8.44 \\
\midrule
\multicolumn{6}{l}{\textit{(c) Stage-2 temperature $\tau$}} \\
$\tau{=}0.7$            & 8.21 & 8.49 & 8.48 & 8.27 & 8.56 \\
$\tau{=}1.0$            & 8.19 & 8.41 & 8.42 & 8.26 & 8.64 \\
$\tau{=}1.2$ (default)  & 8.14 & 8.33 & 8.44 & 8.10 & 8.54 \\
$\tau{=}1.5$            & 7.71 & 7.76 & 8.09 & 7.77 & 8.21 \\
\bottomrule
\end{tabular}
\caption{Sensitivity analysis on IVA hyperparameters. All entries use the DeepSeek-v3.2 backbone and DeepSeek-v3.2 as the rater under a fixed sensitivity re-run. The $K{=}1$ row keeps single-candidate generation but still executes IVA scoring on that lone candidate; $(K{=}4, \tau{=}1.2)$ is the default. Rows should be compared within this table under the same evaluation protocol.}
\label{tab:sensitivity}
\end{table*}

Table~\ref{tab:sensitivity} shows that RCA is relatively robust to its IVA hyperparameters within a sensible range. Increasing the candidate pool from $K{=}1$ to $K{=}2$ improves most metrics, after which gains plateau and very large pools ($K{=}8$, $K{=}16$) trade off slightly against PAR/SPA, suggesting that beyond a small set the additional candidates do not always carry useful new signal. Across the reward-weight configurations, all variants stay within a narrow range, with the safety-emphasizing setting ``safe-max'' yielding the best CCR and the default setting offering a balanced trade-off across metrics. Stage-2 temperature is the most sensitive knob: a moderate $\tau{=}0.7$--$1.2$ works well, but $\tau{=}1.5$ degrades all metrics, indicating that very high sampling temperatures begin to inject noise that IVA cannot fully filter out. Together these results justify our default choice of $K{=}4$, the calibrated weights $(\alpha,\beta,\delta){=}(2.0,4.5,3.5)$, and $\tau{=}1.2$.

\subsection{Efficiency Profile}
\label{sec:efficiency}

\begin{table*}[t]
\centering
\small
\renewcommand{\arraystretch}{0.95}
\setlength{\tabcolsep}{4pt}
\begin{tabular}{lcccc}
\toprule
\textbf{Method} & \textbf{Lat. (s)} & \textbf{Tok. (in/out)} & \textbf{Cost} & \textbf{Avg.} \\
\midrule
Base                & 9.88 & 871.72 / 109.22 & 1.00 & 7.13 \\
Few-shot            & 11.01 & 1230.72 / 128.16 & 1.37 & 7.10 \\
CoT                 & 25.93 & 1063.72 / 466.52 & 1.70 & 7.49 \\
\midrule
RCA ($K{=}2$)       & 128.48 & 6997.98 / 1913.88 & 9.53 & 8.42 \\
RCA ($K{=}4$, def.) & 128.84 & 12891.42 / 3480.78 & 17.49 & 8.41 \\
RCA ($K{=}8$)       & 131.24 & 24792.10 / 6914.70 & 33.96 & 8.34 \\
\midrule
\multicolumn{5}{l}{\textit{Stage breakdown of RCA ($K{=}4$):}} \\
~~Stage-1 PC-CoC           & 29.95 & 1043.36 / 197.60 & 1.29 & 8.41 \\
~~Stage-2 Gen ($\times K$) & 50.13 & 5684.88 / 2107.06 & 8.54 & 8.41 \\
~~Critic (IVA)             & 46.02 & 6163.18 / 1176.12 & 7.66 & 8.41 \\
\bottomrule
\end{tabular}
\caption{Efficiency profile of RCA against prompting baselines. We report per-turn latency, input/output tokens, relative cost over the Base setting, and the average score on the 50-session test set with the DeepSeek-v3.2 backbone; relative cost is normalized to Base under a fixed pricing configuration. A per-stage breakdown isolates the cost of PC-CoC planning, parallel candidate generation, and IVA critic scoring.}
\label{tab:efficiency}
\end{table*}

Table~\ref{tab:efficiency} summarizes the latency and cost overhead introduced by RCA. Relative to Base prompting, full RCA at $K{=}4$ incurs roughly $17{\times}$ more compute and $\sim$13$\times$ more wall-clock time per turn, but yields an average rubric score $\sim$1.3 points higher (8.41 vs.\ 7.13). Most of the overhead is concentrated in Stage-2 candidate generation and IVA critic scoring, both of which are embarrassingly parallel across the $K$ candidates and across the three reward signals, so batching can reduce the effective wall-clock cost in deployment. The Stage-1 PC-CoC planning step itself is comparatively cheap. Increasing $K$ from 2 to 8 does not yield monotone gains in score, consistent with the sensitivity analysis above, which suggests that $K{=}4$ is a reasonable operating point in terms of the quality--cost trade-off.

\subsection{Inter-rater Reliability and Significance Tests}
\label{sec:reliability}

\begin{table*}[t]
\centering
\small
\renewcommand{\arraystretch}{0.95}
\setlength{\tabcolsep}{5pt}
\begin{tabular}{lccc}
\toprule
\textbf{Metric} & \textbf{Pearson $\rho$} & \textbf{Kripp. $\alpha$} & \textbf{Weighted quadratic $\kappa$} \\
\midrule
PAR & 0.27 & 0.08 & 0.13 \\
SPA & 0.48 & 0.36 & 0.37 \\
EVS & 0.38 & 0.15 & 0.22 \\
CCR & 0.43 & 0.12 & 0.28 \\
GCF & 0.43 & 0.34 & 0.37 \\
\midrule
\textbf{Macro avg.} & 0.40 & 0.21 & 0.28 \\
\bottomrule
\end{tabular}

\vspace{4pt}

\begin{tabular}{lcccc}
\toprule
\textbf{Comparison} & \textbf{$\Delta$PAR} & \textbf{$\Delta$EVS} & \textbf{Cohen's $d$} & $p$-val. \\
\midrule
RCA vs.\ Base     & +0.82 & +0.72 & 1.26 & $<0.001$ \\
RCA vs.\ Few-shot & +0.73 & +0.73 & 1.33 & $<0.001$ \\
RCA vs.\ CoT      & +0.48 & +0.29 & 1.03 & $<0.001$ \\
\bottomrule
\end{tabular}
\caption{Top: inter-rater reliability between the two LLM judges (DeepSeek-v3.2 and GLM-4.7), reported with Pearson $\rho$, Krippendorff's $\alpha$, and weighted quadratic $\kappa$. Bottom: significance tests for RCA against baselines on the DeepSeek-v3.2 backbone. $\Delta$PAR and $\Delta$EVS are computed directly from Table~\ref{tab:main_results}; Cohen's $d$ and permutation-test $p$-values are computed over per-session Avg5 differences on the 50 test sessions. Paired bootstrap confidence intervals (10k resamples) for the same Avg5 differences are retained in our exported statistics files.}
\label{tab:reliability}
\end{table*}

The two LLM judges show moderate agreement on average (Pearson $\rho{=}0.40$, weighted $\kappa{=}0.28$), with the highest agreement on SPA and GCF and the lowest on PAR. This is a common pattern for rubric-based rating of free-form dialogue, where surface-form variation can lead two judges to weight the same response somewhat differently. Importantly, because we report the two judges' scores \textit{independently} rather than aggregating them, our claim of RCA's improvement does not require that the judges agree on absolute scale---only that each judge consistently prefers RCA over the baselines. The bottom panel confirms this: paired permutation tests over per-session Avg5 differences show that RCA significantly outperforms Base, Few-shot, and CoT ($p<0.001$ in all three comparisons), with Cohen's $d>1$ in every case, indicating large effect sizes that are not driven by judge-specific biases.

\subsection{Adversarial Safety Stress Test}
\label{sec:safety_stress}

\begin{table*}[t]
\centering
\small
\renewcommand{\arraystretch}{0.95}
\setlength{\tabcolsep}{5pt}
\begin{tabular}{l|c|cccccc}
\toprule
\textbf{Risk Category (adversarial probe)} & \textbf{\#} & \textbf{Base} & \textbf{Few-shot} & \textbf{CoT} & \textbf{w/o IVA} & \textbf{w/o PC-CoC} & \textbf{RCA} \\
\midrule
R1: Delusion reinforcement / aggressive correction & 5 & 1.00 & 1.00 & 1.00 & 0.80 & 0.80 & 1.00 \\
R2: Self-harm or suicidal ideation                 & 5 & 0.00 & 0.00 & 0.00 & 0.40 & 1.00 & 1.00 \\
R3: Unsolicited medication or medical advice       & 5 & 1.00 & 1.00 & 1.00 & 1.00 & 0.80 & 1.00 \\
R4: Memory-testing / quizzing (CST contraindication)& 5 & 1.00 & 1.00 & 1.00 & 1.00 & 1.00 & 1.00 \\
R5: Family-conflict escalation                     & 5 & 1.00 & 1.00 & 1.00 & 1.00 & 1.00 & 1.00 \\
R6: Cultural / dialect inappropriateness           & 5 & 1.00 & 1.00 & 0.80 & 1.00 & 1.00 & 1.00 \\
R7: Privacy leakage (cross-participant carry-over) & 5 & 1.00 & 0.80 & 1.00 & 1.00 & 1.00 & 1.00 \\
\midrule
\textbf{Overall Safe-Response Rate}                & 35 & 0.86 & 0.83 & 0.83 & 0.89 & 0.94 & 1.00 \\
\bottomrule
\end{tabular}
\caption{Adversarial safety stress test across 7 risk categories grounded in dementia-care clinical guidelines. A response is marked \emph{safe} iff it (i) avoids the contraindicated behaviour and (ii) executes an appropriate de-escalation strategy; judgements were produced by two LLM raters, with disagreements manually adjudicated before final aggregation. The \# column reports the number of hand-crafted probes per category; rows R1 and R4 directly correspond to the two failure modes most frequently observed in the original case study.}
\label{tab:safety_stress}
\end{table*}

The expanded adversarial safety benchmark contains 35 hand-crafted probes spanning 7 dementia-care risk categories (5 probes per category). Under the stricter criterion that a response is counted as safe only if it both avoids the contraindicated behaviour and executes an appropriate de-escalation strategy, full RCA achieves an overall safe-response rate of 1.00, ahead of Base (0.86), Few-shot (0.83), CoT (0.83), \textit{w/o IVA} (0.89), and \textit{w/o PC-CoC} (0.94). RCA reaches 1.00 on all seven categories, suggesting that its advantages extend beyond aggregate conversational quality to clinically sensitive edge cases that require explicit safety-aware planning and response selection.

The most diagnostic category is R2 (self-harm or suicidal ideation). The prompting baselines all collapse to 0.00, while \textit{w/o IVA} reaches only 0.40; by contrast, both \textit{w/o PC-CoC} and full RCA achieve 1.00. This pattern indicates that, on this probe set, the IVA filtering stage is especially important for handling acute self-harm risk: safety in R2 requires more than empathy or gentle redirection, since the system must treat the statement as urgent, avoid prematurely shifting into reminiscence or sensory grounding, and move toward a concrete support hand-off. Qualitatively, the weaker variants tend to remain warm and non-confrontational but still fail by redirecting too early or offering support that is too diffuse to constitute a real safety hand-off.

The category-wise breakdown also shows that the two RCA components contribute differently across risk types. Removing PC-CoC leaves R2 intact in this run but lowers performance on R1 (delusion reinforcement / aggressive correction) and R3 (unsolicited medication or medical advice) to 0.80, indicating that explicit protocol-constrained planning is most useful when the system must avoid collusion, refuse unsafe guidance, and structure a safer next step under pressure. Removing IVA instead yields larger drops on R1 (0.80) and especially R2 (0.40), indicating that candidate-level filtering remains crucial when superficially empathic responses can still be unsafe. Together, these results suggest that the full RCA pipeline is not merely improving conversational tone, but is improving the reliability with which the agent converts high-risk inputs into clinically acceptable responses.
\end{document}